%% file: main.tex
\documentclass[10pt,twocolumn,letterpaper]{article}

\usepackage{wacv}              

\input{preamble}

\definecolor{wacvblue}{rgb}{0.21,0.49,0.74}
\usepackage[pagebackref,breaklinks,colorlinks,allcolors=wacvblue]{hyperref}

\def\wacvPaperID{51} 
\def\confName{WACV}
\def\confYear{2027}

\title{ETA-Motion: Efficient Topology-Agnostic \\ Motion Diffusion via Sparse Joint Learning}

\author{
Seoyoung Hwang$^{1,2}$ \quad Tae-Kyun Kim$^{1}$\\
$^{1}$KAIST\\
$^{2}$Samsung Electronics\\
{\tt\small seoyoung11@kaist.ac.kr \qquad kimtaekyun@kaist.ac.kr}
}

\begin{document}
\twocolumn[{
\maketitle
\begin{center}
    \vspace{-1.0em}
    \includegraphics[width=\textwidth]{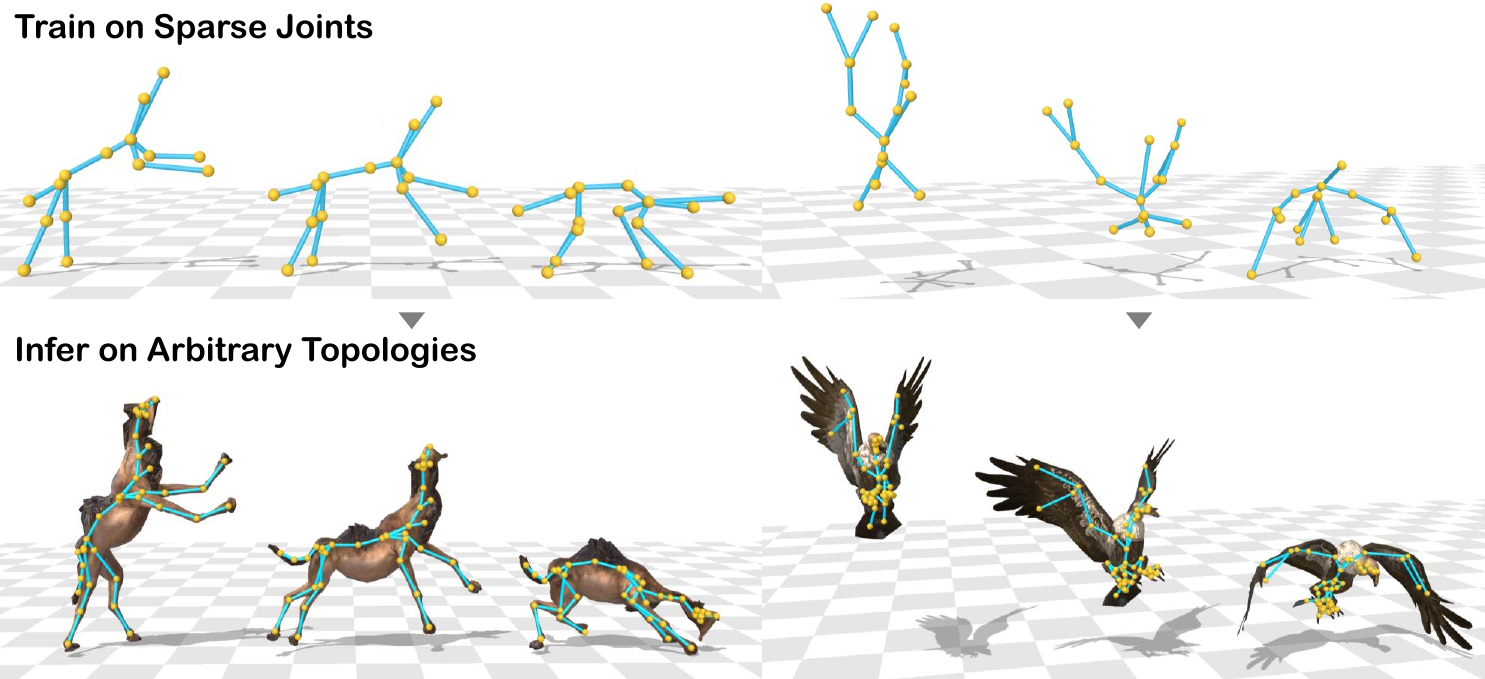}
    \captionof{figure}{ETA-Motion is trained only on sparse joint subsets, yet it can generate motions for full target skeletons at inference time.}
    \label{fig:teaser}
    \vspace{1.0em}
\end{center}
}]
\input{sec/0_abstract}    
\input{sec/1_intro}
\input{sec/2_related_work}
\input{sec/3_method}

\input{sec/4_experiments}
\input{sec/5_conclusion}
{
    \small
    \bibliographystyle{ieeenat_fullname}
    \bibliography{main}
}

\end{document}


\maketitle
\thispagestyle{empty}

\section{Analysis of Joint Selection Heuristics}

In addition to the FPS-based joint selection method presented in the main paper, we compare FPS with connectivity-aware, motion-aware, and random selection. All methods include the root joint and select the same number of joints under the same joint budget \(K\). The connectivity-aware variant applies farthest-point sampling using topological hop distances on the skeleton tree. The motion-aware variant selects the remaining joints with the largest temporal variance in local 6D rotations, while random selection samples the remaining joints uniformly at random.

Table~\ref{tab:joint_selection_heuristics} compares different joint selection heuristics under the same budget \(K=20\). While no alternative consistently outperforms FPS across all settings, random selection achieves the best coverage on unseen topologies, suggesting improved robustness to topology shifts. Together with the joint-budget analysis in Table~\ref{tab:supp_k_budget}, these results suggest that the joint budget has a larger impact than the specific selection heuristic.

\section{Evaluation Metric Details}
\label{sec:supp_metrics}

\subsection{Bone Length Variance (BLV)}
BLV measures the temporal consistency of generated bone lengths. We first recover global joint positions from the motion representation, then compute the length of every skeleton edge at each frame. Let \(E\) denote the set of parent-child edges in the target skeleton and let \(\mathbf{x}_i^t\) be the global 3D position of joint \(i\) at frame \(t\). BLV is defined as
\begin{equation}
\mathrm{BLV} = \frac{1}{|E|} \sum_{(i,j)\in E} \mathrm{Var}_t \left( \lVert \mathbf{x}_i^t - \mathbf{x}_j^t \rVert_2 \right).
\end{equation}
Lower BLV indicates that bone lengths remain more stable over time, reflecting stronger rigidity and better preservation of skeletal structure. BLV is computed only on generated motions.

\subsection{Capsule Collision Ratio (CR)}
CR measures self-collision frequency using a capsule-based geometric proxy. Each bone is modeled as a capsule centered on the line segment connecting a parent-child joint pair. The capsule radius for each bone is set to \(2\%\) of its mean temporal bone length. We consider only non-adjacent bone pairs, namely pairs of bones that do not share a joint.

For each frame and each valid bone pair, we compute the minimum distance between the two bone segments in 3D. A collision is counted when this distance is smaller than the sum of the two capsule radii. CR is then defined as the fraction of colliding \((\text{frame}, \text{bone-pair})\) entries over all tested frames and non-adjacent bone pairs. Lower CR indicates fewer self-collisions and better physical plausibility.

\section{User Study Details}
\label{sec:supp_userstudy}

We provide additional details of the user studies used to evaluate semantic aspects that are not directly captured by the automatic metrics. All studies were conducted with 30 participants. In all user studies, we showed a rest-pose skeleton and an animal image, and displayed the text prompt prominently together with the generated motion results.

\subsection{Baseline Comparison}
For the baseline comparison, participants were shown motion results generated by Ours, AnyTop, and Dragon for the same text prompt and target skeleton. The order of the three methods was randomized for each question. For each example, participants were asked to select the single best result for the following two questions: \textit{skeleton-motion compatibility}: ``Which motion is the most natural and appropriate for the given skeleton?'' and \textit{Text-motion alignment}: ``Which motion best matches the given text description?'' We report the preference score of each method as the percentage of responses in which it was selected as the best result.

\subsection{Joint Budget Comparison}
For the joint budget comparison, participants were shown pairs of motion results generated with different joint budgets \(K \in \{5, 10, 20, 50, \text{full}\}\). For each pair, the order of presentation was randomized, and participants selected the preferred result using the same two questions as in the baseline comparison: \textit{skeleton--motion compatibility}: ``Which motion is the most natural and appropriate for the given skeleton?'' and \textit{text--motion alignment}: ``Which motion best matches the given text description?'' We aggregate the pairwise comparison results using the Bradley--Terry model, which converts pairwise wins and losses into a global preference score for each joint-budget setting, and report the resulting scores in the main paper.

\section{Detailed Experimental Analysis}
\label{sec:supp_experiments}

\subsection{Ablation Details}
Table~\ref{tab:supp_k_budget} provides the full quantitative results for the joint budget ablation corresponding to the graph in the main paper. Table~\ref{tab:component_ablation} provides the full metric breakdown for the component ablations summarized in the main paper. The results support the main-paper observation that increasing \(K\) improves performance only up to a moderate budget. In particular, \(K=20\) achieves the best Coverage across all three settings while remaining substantially more efficient than larger budgets and the full-joint setting. Although \(K=50\) remains competitive, it does not translate into better overall motion quality or topology-agnostic generalization than the proposed \(K=20\) setting.

\subsection{Effect of Visual Joint Embedding}

We further evaluate the robustness of the visual joint embedding on the pig asset used in the main paper, whose joints are labeled with arbitrary identifiers (e.g., \texttt{joint1}, \texttt{joint2}) rather than semantic names. We compare motion samples generated by models trained with and without the visual joint embedding using the same text prompt: \textit{``An animal runs energetically forward.''}

\begin{figure}[h]
    \centering
    \includegraphics[width=\columnwidth]{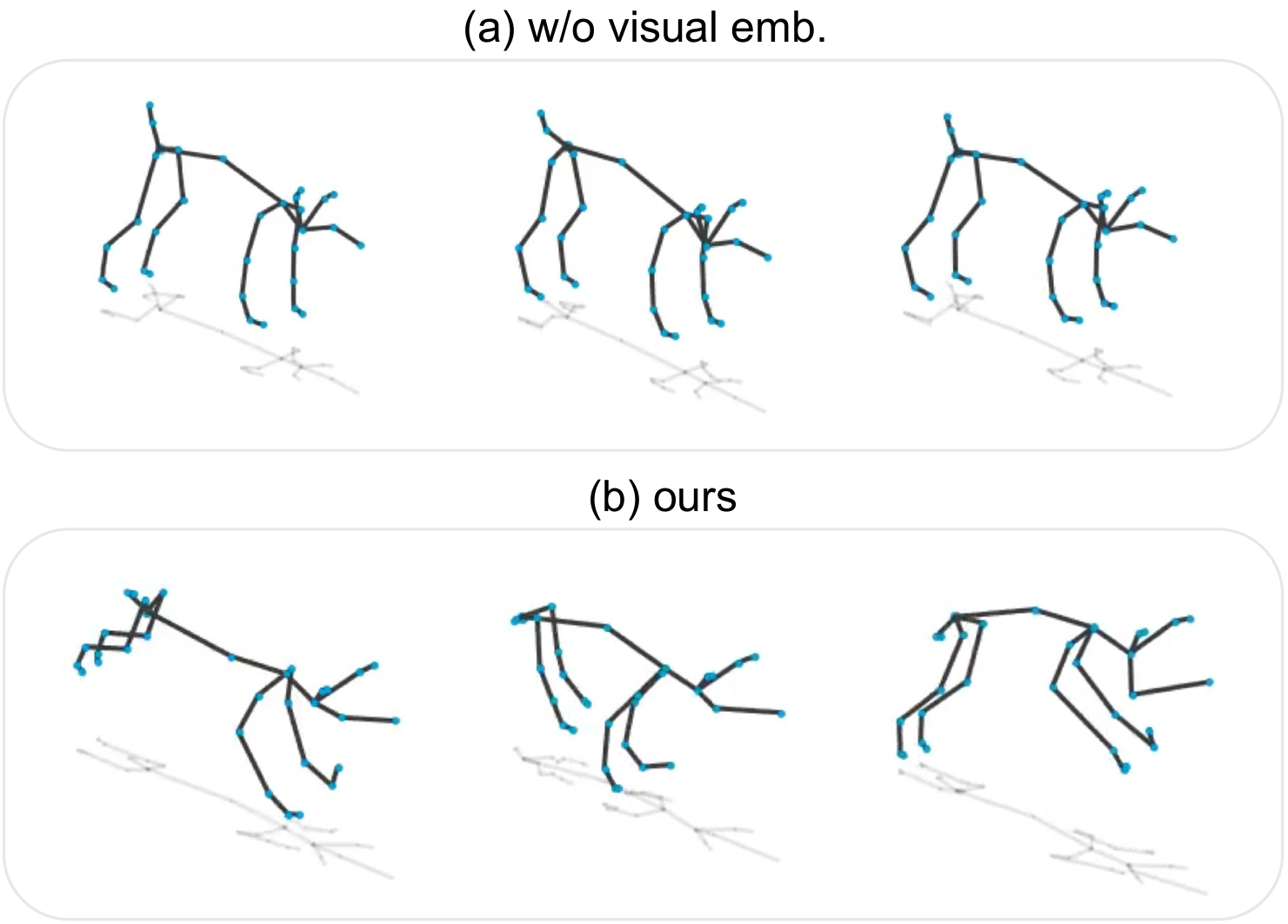}
    \caption{Qualitative comparison under noisy joint labels. With arbitrary joint identifiers, the model trained with the visual joint embedding generates a forward quadrupedal running motion, whereas the model trained without it produces only subtle motion.}
    \label{fig:noisy_label_qualitative}
\end{figure}

As shown in Figure~\ref{fig:noisy_label_qualitative}, the model trained without the visual joint embedding produces only subtle overall motion and fails to generate a running motion. In contrast, the model trained with the visual joint embedding generates a plausible quadrupedal running motion. These results demonstrate that visual joint embeddings improve robustness to uninformative joint labels and facilitate practical motion generation for rigged assets with noisy joint annotations.

\section{Qualitative Animation Results}
\label{sec:supp_video_mesh}

We provide a supplementary video containing qualitative comparisons with
baseline methods, mesh animation results, and additional motion examples.
As motion quality and temporal consistency are best evaluated dynamically,
the video provides a clearer view of the generated motions than static
figures alone. All models shown in the video are trained for 600K steps
under the same setting as the main comparison.

\begin{figure*}[t]
    \centering

    \begin{subfigure}[t]{\textwidth}
        \centering
        \includegraphics[width=\linewidth]{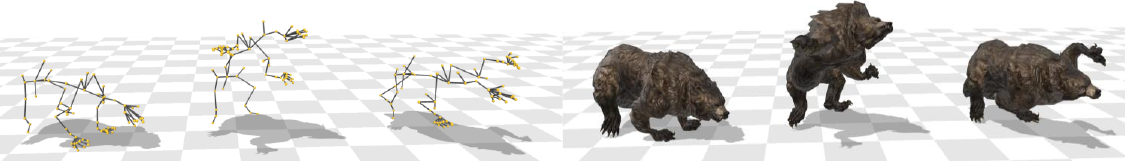}
        \caption{\small\textit{A bear swings its powerful front paw to assert dominance.}}
    \end{subfigure}

    \vspace{3pt}

    \begin{subfigure}[t]{\textwidth}
        \centering
        \includegraphics[width=\linewidth]{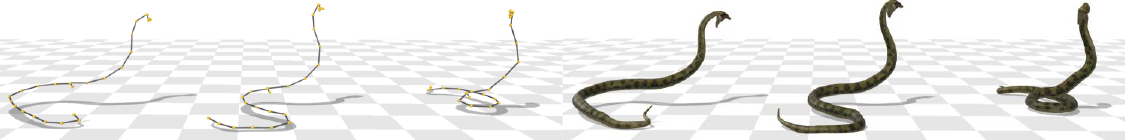}
        \caption{\small\textit{An anaconda raises its head and neck while curling its tail inward.}}
    \end{subfigure}

    \vspace{3pt}

    \begin{subfigure}[t]{\textwidth}
        \centering
        \includegraphics[width=\linewidth]{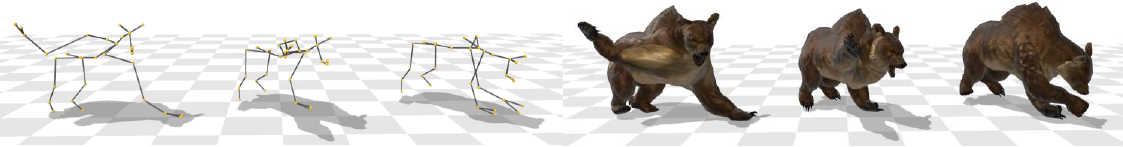}
        \caption{\small\textit{A quadruped stands on all fours, rears up, and lunges forward to bite with its powerful jaws.}}
    \end{subfigure}

    \vspace{3pt}

    \begin{subfigure}[t]{\textwidth}
        \centering
        \includegraphics[width=\linewidth]{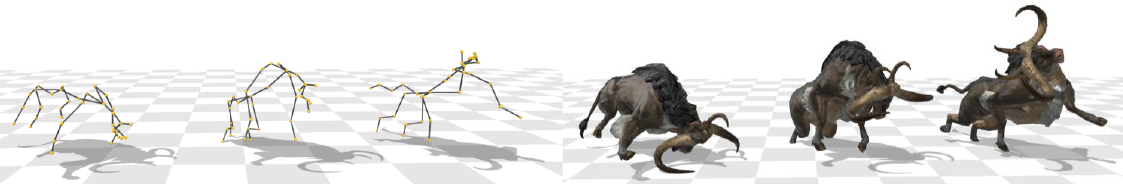}
        \caption{\small\textit{A buffalo attacks by lowering its body and thrusting its head upward.}}
    \end{subfigure}

    \vspace{3pt}

    \begin{subfigure}[t]{\textwidth}
        \centering
        \includegraphics[width=\linewidth]{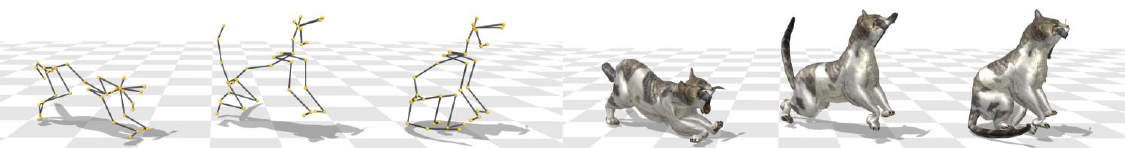}
        \caption{\small\textit{A cat stretches and yawns, then returns to a sitting position.}}
    \end{subfigure}

    \vspace{3pt}

    \begin{subfigure}[t]{\textwidth}
        \centering
        \includegraphics[width=\linewidth]{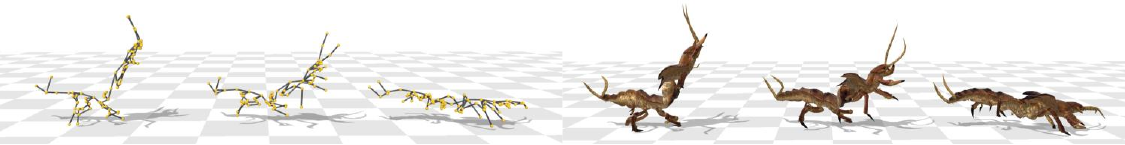}
        \caption{\small\textit{A centipede prepares and executes an attack by raising its body and striking with its arms.}}
    \end{subfigure}

    \vspace{3pt}

    \begin{subfigure}[t]{\textwidth}
        \centering
        \includegraphics[width=\linewidth]{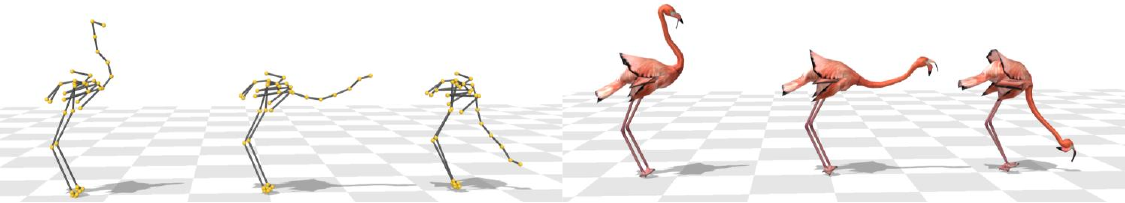}
        \caption{\small\textit{A flamingo stands still, bending its neck to peck at the ground.}}
    \end{subfigure}

    \caption{Mesh animation results across diverse skeleton topologies.
    Each subfigure shows the inverse-kinematics-processed joint motion on
    the left and the resulting character mesh animation on the right.}
    \label{fig:supp_mesh}
\end{figure*}

To visualize the generated motions on articulated characters, we recover
joint rotations from the generated joint positions using inverse kinematics.
We then apply the recovered rotations to the original character meshes
using linear blend skinning with the skinning weights provided by their
rigs. We neither recompute the skinning weights nor re-rig the meshes.
Figure~\ref{fig:supp_mesh} presents representative mesh animation
results, while the supplementary video shows the corresponding temporal
motions.
\begin{table*}[t]
    \centering
    \scriptsize
    \setlength{\tabcolsep}{4pt}
    \renewcommand{\arraystretch}{1.1}
    \caption{Comparison of alternative joint selection heuristics under a joint budget of \(K=20\).}
    \label{tab:joint_selection_heuristics}
    \resizebox{\textwidth}{!}{%
    \begin{tabular}{lccccccccc}
        \toprule
        \multirow{2}{*}{Heuristic}
        & \multicolumn{3}{c}{Seen}
        & \multicolumn{3}{c}{Unseen Motion}
        & \multicolumn{3}{c}{Unseen Topology} \\
        \cmidrule(lr){2-4} \cmidrule(lr){5-7} \cmidrule(lr){8-10}
        & Coverage $\uparrow$ & BLV $\downarrow$ & CR $\downarrow$
        & Coverage $\uparrow$ & BLV $\downarrow$ & CR $\downarrow$
        & Coverage $\uparrow$ & BLV $\downarrow$ & CR $\downarrow$ \\
        \midrule
        FPS & 88.60 & 0.0013 & 0.0009 & 75.57 & 0.0009 & 0.0009 & 65.73 & 0.0010 & 0.0008 \\
        Random & 86.23 & 0.0010 & 0.0010 & 75.30 & 0.0007 & 0.0008 & 68.55 & 0.0006 & 0.0007 \\
        Connectivity-aware & 80.47 & 0.0017 & 0.0009 & 74.23 & 0.0010 & 0.0010 & 64.64 & 0.0009 & 0.0009 \\
        Motion-aware & 83.02 & 0.0014 & 0.0011 & 73.64 & 0.0009 & 0.0011 & 62.51 & 0.0008 & 0.0009 \\
        \bottomrule
    \end{tabular}%
    }
\end{table*}

\begin{table*}[t]
    \centering
    \setlength{\tabcolsep}{5.5pt}
    \renewcommand{\arraystretch}{1.15}
    \caption{Full results for the joint budget ablation at 300K training steps. All variants use the proposed position-only motion representation. We report Coverage, BLV, and CR for the Seen, Unseen Motion, and Unseen Topology settings. The row ``full'' corresponds to the full-joint setting. Best results are highlighted in bold.}
    \label{tab:supp_k_budget}
    \begin{tabular*}{\textwidth}{@{\extracolsep{\fill}} c ccc ccc ccc}
        \toprule
        \multirow{2}{*}{\(K\)}
        & \multicolumn{3}{c}{Seen}
        & \multicolumn{3}{c}{Unseen Motion}
        & \multicolumn{3}{c}{Unseen Topology} \\
        \cmidrule(lr){2-4} \cmidrule(lr){5-7} \cmidrule(lr){8-10}
        & Coverage $\uparrow$ & BLV $\downarrow$ & CR $\downarrow$
        & Coverage $\uparrow$ & BLV $\downarrow$ & CR $\downarrow$
        & Coverage $\uparrow$ & BLV $\downarrow$ & CR $\downarrow$ \\
        \midrule
        5
        & 44.16 & 0.0195 & 0.0025
        & 32.99 & 0.0136 & 0.0023
        & 30.54 & 0.0127 & 0.0031 \\
        10
        & 77.80 & 0.0059 & 0.0016
        & 65.87 & 0.0038 & 0.0012
        & 52.94 & 0.0050 & 0.0017 \\
        20
        & \textbf{88.60} & \textbf{0.0013} & \textbf{0.0009}
        & \textbf{75.57} & \textbf{0.0009} & 0.0009
        & \textbf{65.73} & \textbf{0.0010} & \textbf{0.0008} \\
        50
        & 74.06 & 0.0033 & 0.0010
        & 66.49 & 0.0019 & \textbf{0.0008}
        & 53.36 & 0.0018 & 0.0009 \\
        full
        & 67.46 & 0.0047 & 0.0015
        & 56.50 & 0.0035 & 0.0013
        & 42.45 & 0.0044 & 0.0016 \\
        \bottomrule
    \end{tabular*}
\end{table*}

\begin{table*}[t]
    \centering
    \scriptsize
    \setlength{\tabcolsep}{4pt}
    \renewcommand{\arraystretch}{1.12}
    \caption{Component ablations of ETA-Motion at 300K training steps. We report Coverage, BLV, and CR on the Seen, Unseen Motion, and Unseen Topology settings. Best results are highlighted in bold within each block.}
    \label{tab:component_ablation}
    \resizebox{\textwidth}{!}{%
    \begin{tabular}{ll ccc ccc ccc}
        \toprule
        \multirow{2}{*}{K} & \multirow{2}{*}{Method}
        & \multicolumn{3}{c}{Seen}
        & \multicolumn{3}{c}{Unseen Motion}
        & \multicolumn{3}{c}{Unseen Topology} \\
        \cmidrule(lr){3-5} \cmidrule(lr){6-8} \cmidrule(lr){9-11}
        & 
        & Coverage $\uparrow$ & BLV $\downarrow$ & CR $\downarrow$
        & Coverage $\uparrow$ & BLV $\downarrow$ & CR $\downarrow$
        & Coverage $\uparrow$ & BLV $\downarrow$ & CR $\downarrow$ \\
        \midrule
        \multirow{3}{*}{20}
        & Ours
        & \textbf{88.60} & \textbf{0.0013} & \textbf{0.0009}
        & \textbf{75.57} & \textbf{0.0009} & \textbf{0.0009}
        & 65.73 & \textbf{0.0010} & \textbf{0.0008} \\
        & w/o visual emb.
        & 82.52 & 0.0025 & 0.0022
        & 73.15 & 0.0016 & 0.0022
        & \textbf{68.41} & 0.0022 & 0.0019 \\
        & w/o bone loss
        & 77.52 & 0.0026 & 0.0012
        & 67.50 & 0.0018 & 0.0012
        & 63.74 & 0.0021 & 0.0012 \\
        \midrule
        \multirow{2}{*}{Full}
        & Ours
        & \textbf{67.46} & \textbf{0.0047} & \textbf{0.0015}
        & \textbf{56.50} & \textbf{0.0035} & \textbf{0.0013}
        & \textbf{42.45} & \textbf{0.0044} & \textbf{0.0016} \\
        & w/ redundant rep.
        & 50.34 & 0.1755 & 0.0041
        & 48.03 & 0.0560 & 0.0024
        & 35.29 & 0.0397 & 0.0039 \\
        \bottomrule
    \end{tabular}%
    }
\end{table*}

%% file: preamble.tex
\usepackage{booktabs}
\usepackage{multirow}
\usepackage{graphicx}
\usepackage{makecell}
\usepackage{subcaption}
\usepackage[percent]{overpic}
\usepackage{comment}
\usepackage{pgfplots}
\pgfplotsset{compat=1.18}

%% file: sec/0_abstract.tex
\begin{abstract}
Topology-agnostic motion generation aims to synthesize motions for diverse characters with arbitrary skeletons. Unlike conventional motion generation, which assumes a fixed skeletal template, this setting must handle diverse joint configurations and inconsistent joint semantics across characters. Existing methods address this challenge through dense full-joint modeling, which incurs substantial computational cost and hinders learning transferable motion patterns across diverse skeletal topologies. We present ETA-Motion, a topology-agnostic motion generation framework based on sparse joint learning. Instead of learning motion from all joints, our method uses a compact subset of joints to reduce redundancy and trains the motion generator on the resulting sparse skeleton. To facilitate correspondence across diverse topologies, we introduce semantic joint embeddings that combine textual and visual cues, together with topology-aware training objectives for preserving skeletal structure. Although ETA-Motion is trained only on sparse skeletal observations, it supports full-skeleton motion generation during inference. Experimental results demonstrate that ETA-Motion outperforms existing topology-agnostic baselines while significantly improving efficiency and generalization to unseen motions and unseen topologies.
\end{abstract}

%% file: sec/1_intro.tex
\section{Introduction}
\label{sec:intro}

Recent advances in generative modeling have significantly improved human motion generation. A wide range of approaches, including autoregressive modeling, masked modeling, and diffusion models, have enabled the synthesis of realistic and controllable motions under diverse conditions~\cite{Zhang23T2MGPT,Guo24MoMask,Pinyoanuntapong24MMM, jiang2024motiongpt, guo2022tm2t, zhong2023attt2m, pinyoanuntapong2024bamm, Zhang22MotionDiffuse,Chen23Executing,Tevet23MDM,Zhang23ReMoDiffuse, wang2023fg, dai2024motionlcm, zhang2024motion}. Most of these methods, however, are developed under a fixed-template setting, where all training samples share the same skeletal structure, joint ordering, and kinematic hierarchy. Under this assumption, joint semantics are implicitly aligned across the dataset~\cite{Guo2022HumanML3D}, and modeling skeletal attributes such as joint identity or topology is rarely treated as a primary challenge.

Topology-agnostic motion generation aims to directly synthesize motions conditioned on an arbitrary input skeleton, without relying on a fixed skeletal template. This setting is substantially more challenging because joint indices no longer carry consistent semantics across samples, and the underlying articulation structures vary significantly across characters. Recent works such as AnyTop~\cite{Gat25AnyTop} and How to Move Your Dragon~\cite{Lee25Dragon} take important steps toward this setting by learning a single model over diverse skeletons while preserving full joint dimensionality during motion modeling.

Despite this progress, dense full-joint modeling introduces two fundamental limitations. First, it incurs substantial computational overhead as the number of joints increases. More importantly, maintaining the full joint dimensionality throughout training encourages the model to learn topology-specific details that are not always necessary for transferable motion generation. As a result, the complexity of topology-agnostic motion learning is not merely an efficiency issue, but is also closely tied to motion quality and generalization.

To address this challenge, we present ETA-Motion, a topology-agnostic motion generation framework based on sparse joint learning. Instead of training on full-joint motion data, ETA-Motion learns motion dynamics from a compact subset of joints, reducing redundancy in motion modeling, while improving computational efficiency. To support correspondence across diverse skeletons, we further introduce semantic joint embeddings that combine textual and visual cues, together with topology-aware objectives for preserving structural consistency. As illustrated in Fig.~\ref{fig:teaser}, although the model is trained on sparse joint subsets, it generates motions for target skeletons with arbitrary topologies at inference time. Our main contributions are as follows.
\begin{itemize}
\item We propose a sparse joint subset learning framework for topology-agnostic motion generation, which replaces dense full-joint training with a compact sparse representation to reduce redundancy and computational cost.
\item We design a topology-aware motion learning scheme with semantic joint embeddings and structural objectives, enabling transferable motion generation across diverse skeleton topologies while preserving skeletal consistency.
\item Extensive experiments show that training on sparse joint subsets improves not only computational efficiency, but also motion quality and topology preservation over existing topology-agnostic baselines.
\end{itemize}

%% file: sec/2_related_work.tex
\section{Related Work}
\label{sec:related}

\begin{figure*}[t]
    \centering
    \includegraphics[width=\textwidth]{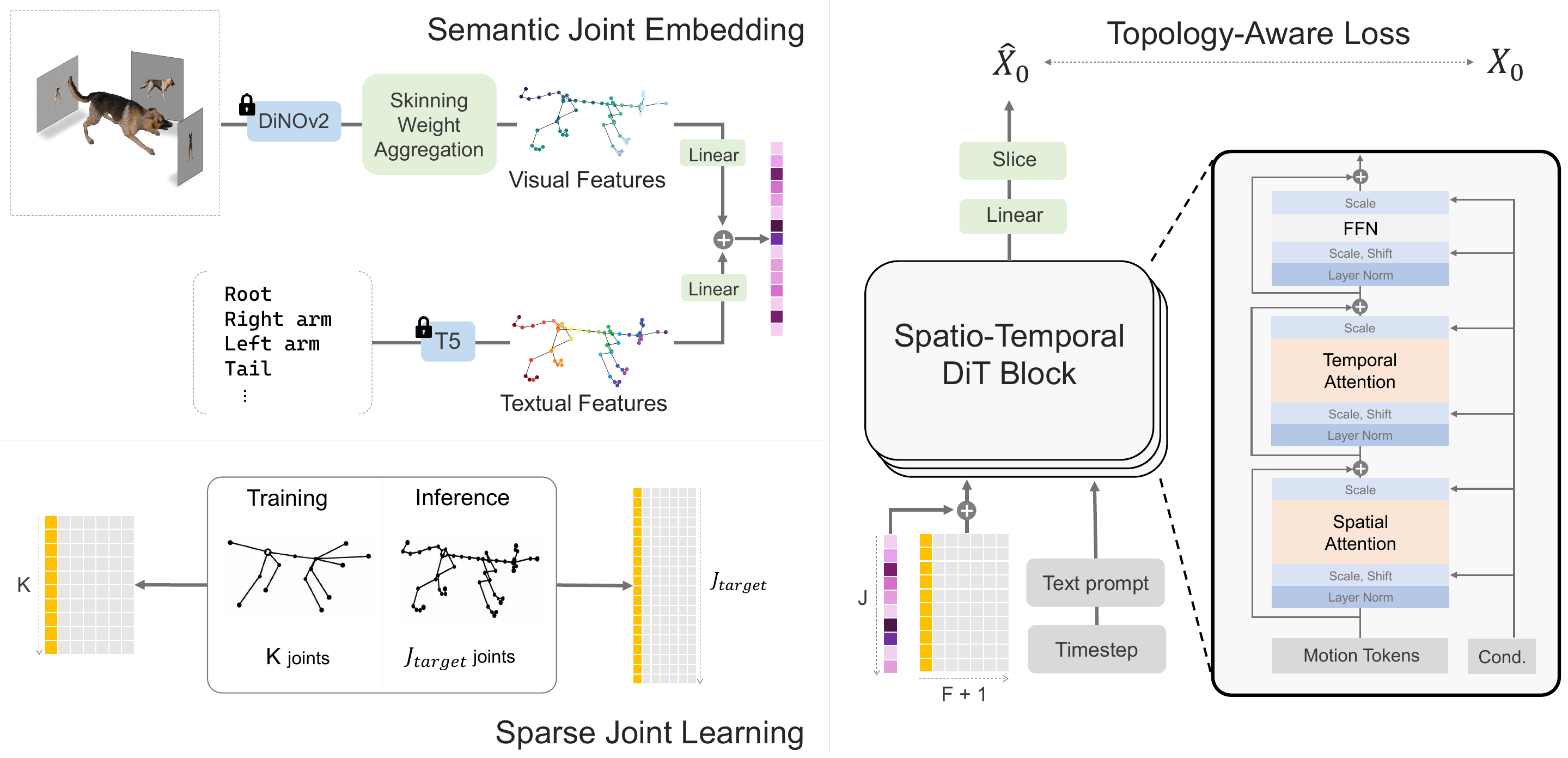}
    \caption{Overview of ETA-Motion. The model is trained on sparse joint subsets but generates motions for full target skeletons at inference time. To enable transferable motion generation across diverse topologies, we combine sparse joint learning, semantic joint embeddings from textual and visual cues, and topology-aware supervision within a spatio-temporal DiT framework.}
    \label{fig:method_overview}
\end{figure*}

\subsection{Human Motion Generation}
Human motion generation has advanced substantially with the development of modern generative models. Existing approaches can be broadly categorized into two major paradigms: discrete token-based methods and diffusion-based methods. The first line of work represents human motion as compact discrete tokens, typically through vector quantization, and performs motion generation in the token space \cite{Zhang23T2MGPT,Guo24MoMask,Pinyoanuntapong24MMM, jiang2024motiongpt, guo2022tm2t, zhong2023attt2m, pinyoanuntapong2024bamm}. These approaches enable efficient and stable motion generation through compact discrete representations. The second line of work employs diffusion models to model human motion through denoising in continuous space \cite{Zhang22MotionDiffuse,Chen23Executing,Tevet23MDM,Zhang23ReMoDiffuse, wang2023fg, dai2024motionlcm, zhang2024motion, lee25REWIND}. Recent work extends motion diffusion to physics-based control by blending pretrained controllers through reinforcement learning~\cite{nam26dynamichoi}. These methods have demonstrated strong performance in producing realistic and diverse human motions. Despite their different generation mechanisms, both lines of work implicitly assume a fixed human skeletal template, where joint numbers, joint ordering, and kinematic relations are predefined and shared across all samples. As a result, individual joints are rarely treated as explicit modeling units, and skeletal topology is treated as a predefined condition rather than a modeling variable.

In addition, HumanML3D~\cite{Guo2022HumanML3D} has largely standardized the motion representation used in modern motion generation frameworks. Following its formulation, motion is typically represented as a combination of joint positions, relative rotations, local velocities, and foot-contact labels. This design explicitly employs a redundant motion representation, where multiple kinematic attributes are jointly modeled. Consequently, MDM~\cite{Tevet23MDM} introduces auxiliary kinematic objectives, such as geodesic rotation losses and foot-sliding penalties, to maintain consistency among these motion features.

\subsection{Topology-Agnostic Motion Generation}
Generating motions across diverse skeletal topologies remains challenging due to the limited availability of large-scale topology-diverse motion datasets. To alleviate this issue, several works~\cite{Huang25AniMax,Song25Puppeteer} leverage strong priors from large video generation models. While these approaches benefit from large-scale video knowledge, they rely on multi-stage pipelines involving intermediate video generation, leading to increased computational cost and error accumulation.

Recent efforts have begun to directly address topology-aware motion generation without relying on large pretrained priors. AniMo~\cite{Wang25AniMo} adopts VQ-VAE-based discrete motion tokenization and performs species-aware motion generation in the token space. However, it aligns motions from different species into a shared skeletal representation with consistent joint ordering and dimensionality, limiting its ability to model a wide scope of skeletal topology changes. Diffusion-based approaches have emerged as a promising direction for topology-aware motion generation~\cite{Gat25AnyTop,Lee25Dragon}. By incorporating skeletal-aware attention mechanisms, these approaches enable a single model to operate on diverse skeletal topologies. 

However, the increased modeling complexity makes it challenging to learn meaningful motion dynamics that generalize across diverse skeletons. Furthermore, inheriting redundant motion representations from human motion generation significantly increases the complexity of the learning problem, requiring the model to simultaneously capture multiple correlated kinematic attributes. To make learning tractable, prior works often rely on additional dataset-dependent priors, such as extensive topology augmentation~\cite{Lee25Dragon} or joint-wise normalization statistics computed separately for individual characters~\cite{Gat25AnyTop}. Although these designs facilitate optimization, they also increase dependency on object-specific preprocessing and motion priors, making it harder to learn a fully shared motion space that generalizes across diverse topologies.

\subsection{Efficient Motion Generation}
Recent studies show that reducing redundancy in motion representations improves not only computational efficiency but also generation quality. Several works have questioned the necessity of the redundant kinematic representation popularized by HumanML3D, demonstrating that simplified motion representations improve generation quality while reducing modeling complexity \cite{Meng25AbsoluteCoordinates,meng2025rethinking}. Similarly, Less is More~\cite{Bae25LessMore} observes that dense motion sequences contain many redundant frames and proposes a sparse keyframe diffusion framework, leading to both improved motion quality and reduced computational cost.

Several works further explore compact motion representations to improve efficiency. SALAD~\cite{Hong25SALAD} introduces a skeleton-aware latent representation that compresses both temporal and skeletal dimensions, retaining only a small set of atomic joints within the latent space while preserving motion dynamics. SFControl~\cite{hwang2025motion} employs a two-stage framework that first synthesizes sparse key-joint trajectories and subsequently reconstructs full-body motions, demonstrating that low-dimensional skeletal representations effectively support controllable motion synthesis.

Motivated by these findings, we investigate sparse joint learning to reduce skeletal redundancy and enable efficient motion generation across diverse topologies.

%% file: sec/3_method.tex
\section{Method}
\label{sec:method}

\subsection{Preliminaries}
Each asset in the dataset is represented as
$\mathcal{A} = (\mathcal{J}, \mathcal{E}, \mathcal{M})$, where the joint set
\(
\mathcal{J}=(\mathbf{N},\mathbf{P})
\)
contains semantic joint names
\(
\mathbf{N}
\)
and rest-pose
\(
\mathbf{P}
\).
The edge set
\(
\mathcal{E}
\)
consists of directed parent-child connections that define the kinematic hierarchy of the skeleton.
The rigged mesh
\(
\mathcal{M}=(\mathcal{V}, \mathbf{W})
\)
consists of mesh vertices
\(
\mathcal{V}
\)
and skinning weights
\(
\mathbf{W}
\),
where \(W_{v,i}\) denotes the influence weight of joint \(i\) on vertex \(v\).

Given a skeletal motion sequence, the mesh is deformed according to the motion of the underlying joints and the predefined skinning weights. Therefore, our generator does not directly model mesh deformation; instead, it predicts skeletal motion features, from which the posed mesh is obtained through the skinning function.

\subsection{Topology-Agnostic Motion Representation}

Motion representation is particularly important for topology-agnostic motion generation. Representations that are highly tied to a specific kinematic chain, such as joint rotations or parent-relative joint offsets, encourage the model to learn motion features with skeleton-specific meanings. While this is acceptable in fixed-template motion generation, where all samples share the same skeletal structure, it becomes problematic in topology-agnostic motion generation, where the input skeleton itself varies across samples. Consequently, such representations make it difficult to learn transferable motion features across diverse topologies and often lead to performance degradation.

To avoid this issue, we directly model 3D joint positions. Inspired by MARDM~\cite{meng2025rethinking} and
ACMDM~\cite{Meng25AbsoluteCoordinates}, we represent a motion sequence
using 3D joint positions, denoted by
\(X \in \mathbb{R}^{F \times J \times 3}\), where \(F\) and \(J\) denote
the numbers of frames and joints, respectively. Both \(F\) and \(J\) may
vary depending on the motion sequence and input skeleton. Specifically, we represent the root joint in global coordinates and all non-root joints relative to the root joint. This position-based representation allows the model to represent motion without relying on skeleton-specific kinematic parameterizations.

\subsection{Architecture}
Our model is a conditional diffusion model that generates motion sequences from a text description and an arbitrary skeleton. As illustrated in Fig.~\ref{fig:method_overview}, our model adopts a Diffusion Transformer (DiT) backbone with AdaLN-Zero conditioning~\cite{peebles2023scalable}. Given a noisy motion sample \(X_t\), the network predicts the corresponding clean motion \(X_0\).

Before being processed by the DiT backbone, the input motion sequence is transformed into motion tokens through conditioning and embedding. Following AnyTop~\cite{Gat25AnyTop}, we prepend the clean rest pose \(\mathbf{P} \in \mathbb{R}^{J \times 3}\) to the noisy motion sequence \(X_t\). This prepended rest pose serves as an explicit structural reference, providing a clear signal of the target skeleton that should be preserved throughout generation. The transformer therefore receives \((F+1)\times J\) tokens. When all joints are used during training, \(J\) corresponds to \(J_{\max}\), the maximum number of joints in the training set. Skeletons with fewer than \(J_{\max}\) joints are padded, and the padded joints are ignored using a validity mask. During inference, \(J\) corresponds to \(J_{\text{target}}\), the actual number of joints of the target skeleton, so no padding is required. Since the DiT backbone operates on token sequences, it supports varying token lengths induced by different numbers of joints.

The DiT backbone consists of alternating spatial attention, temporal attention, and feed-forward network blocks. Spatial attention models interactions among joints within each frame, while temporal attention captures motion dynamics across frames for each joint. The validity mask is applied in the spatial attention blocks so that padded joints neither attend to other joints nor are attended to. Temporal attention is computed independently for each joint across frames, and padded joint slots are excluded from supervision. The computational complexity of spatial attention scales as \(O(FJ^2)\), whereas that of temporal attention scales as \(O(JF^2)\). Consequently, using all joints during training leads to a rapidly increasing attention cost as \(J_{\max}\) grows. This observation motivates our sparse joint learning.

\subsection{Sparse Joint Learning}

\begin{figure}[t]
    \centering
    \includegraphics[width=\columnwidth]{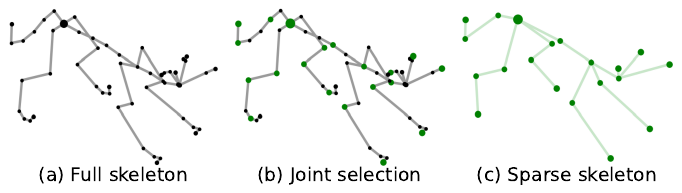}
    \caption{
    Visualization of sparse joint skeletons. From the (a) original skeleton, (b) a compact set of joints is retained, and (c) the retained joints are reconnected to form a sparse yet topology-consistent skeleton.
    }
    \label{fig:key_joint_selection}
\end{figure}

Motivated by the quadratic complexity of spatial attention, ETA-Motion adopts a sparse joint subset training strategy. While this design is primarily introduced for computational efficiency, it is also motivated by an important observation: dense skeletal motions contain substantial redundancy. Neighboring joints within the same articulated part often follow highly correlated motion patterns, while their relative geometric relationships are already determined by the reference skeleton. Consequently, learning motion dynamics from every joint is often unnecessary, and a representative subset of joints is sufficient to capture the underlying motion dynamics.

Accordingly, we train the motion generator using a compact subset of joints under a joint budget of $K$. Since $K \ll J_{\max}$, the spatial attention complexity during training is reduced from $O(FJ_{\max}^2)$ to $O(FK^2)$, substantially reducing both computation and memory usage while also minimizing unnecessary padding in batched training.

Given an input skeleton, we construct a sparse skeleton by selecting \(K\) joints from the original skeleton, as shown in Fig.~\ref{fig:key_joint_selection}. While multiple joint selection strategies are possible, we adopt farthest point sampling (FPS) starting from the root joint as a simple and deterministic heuristic. We then redefine the skeletal connectivity by reconnecting each retained joint to its nearest retained ancestor in the original kinematic hierarchy, yielding a sparse yet structurally consistent skeleton. Additional comparisons with alternative joint selection heuristics are provided in the supplementary material.

During training, the motion generator observes only sparse skeletons. Yet, this sparse training setup generalizes to full target skeletons at inference time. The clean rest pose $\mathbf{P}$ provides an explicit structural signal for the target skeleton, allowing the learned motion dynamics to transfer to full skeletons with different joint counts.

\subsection{Semantic Joint Embedding}

A key challenge in topology-agnostic motion generation is that joint indices do not carry consistent semantics across different skeletons. The same index often corresponds to different anatomical structures depending on the target topology, making it difficult for the model to establish correspondence across skeletons and learn transferable motion patterns. Therefore, explicit semantic encoding is essential for topology-agnostic motion generation.

To this end, we construct a semantic joint embedding by combining textual and visual cues for each joint. Similar to AnyTop~\cite{Gat25AnyTop}, we use joint names as textual cues, and further augment them with visual cues derived from the rigged mesh. Let \(H \in \mathbb{R}^{(F+1)\times J \times d}\) denote the motion tokens obtained by concatenating the projected rest pose and noisy motion sequence, where each token is a \(d\)-dimensional feature vector. We compute the semantic joint embedding as
$E = E^{\text{text}} + E^{\text{vis}}$,
where \(E^{\text{text}} \in \mathbb{R}^{J \times d}\) and \(E^{\text{vis}} \in \mathbb{R}^{J \times d}\) denote the textual and visual joint embeddings, respectively. We add \(E\) to the motion tokens along the joint dimension while broadcasting it across the temporal dimension, yielding \(H' = H + E\). The resulting tokens \(H'\) are then passed to the transformer backbone.

\paragraph{Textual Joint Embedding.}
We derive the textual joint embedding from the joint names \(\mathbf{N}\) defined in the asset representation. These joint names are encoded using a pretrained T5 text encoder and projected into the motion latent space through a linear layer to obtain the textual joint embedding \(E^{\text{text}}\). However, textual cues alone are insufficient for establishing accurate correspondence across diverse skeletons. Joints with the same textual label often exhibit substantially different visual characteristics and motion roles depending on the target topology. For example, joints labeled as \texttt{arm} correspond to forelimbs of quadrupeds or wings of flying creatures in different skeletons. Moreover, asset-provided joint names are often noisy, incomplete, or inconsistent across rigs. These limitations make it difficult to establish accurate correspondence using textual cues alone.

\paragraph{Visual Joint Embedding.}
To provide complementary semantic cues, we construct a visual joint embedding from the rigged mesh. Specifically, we first compute a feature \(f_v\) for each mesh vertex \(v \in \mathcal{V}\) by aggregating DINOv2~\cite{oquab2024dinov2} features from multiple rendered views:
\begin{equation}
f_v
=
\frac{
\sum_{i=1}^{N_{\mathrm{view}}}
M_i(v)\, F^{(i)}(\pi_i(v))
}{
\sum_{i=1}^{N_{\mathrm{view}}}
M_i(v)
},
\end{equation}
where \(M_i(v)\in\{0,1\}\) indicates whether vertex \(v\) is visible in the \(i\)-th rendered view, \(F^{(i)}\) denotes the corresponding DINOv2 feature map, and \(\pi_i(v)\) denotes the projected pixel location of vertex \(v\) in that view. We then aggregate the vertex-level features into joint-level features using the skinning weights:
\begin{equation}
f_j
=
\frac{
\sum_{v\in\mathcal{V}} W_{v,j}\, f_v
}{
\sum_{v\in\mathcal{V}} W_{v,j}
},
\end{equation}
where \(W_{v,j}\) denotes the skinning weight of vertex \(v\) for joint \(j\). Finally, the resulting joint-level features are projected into the motion latent space to obtain the visual joint embedding \(E^{\text{vis}}\).

\subsection{Topology-Aware Training Objective}

Given the text prompt and target skeleton, our model is trained to predict the clean motion sequence $X_0$ from a noisy input. Let $\hat{X}_0$ denote the predicted motion. We employ a reconstruction loss on the selected joints:

\begin{equation}
\mathcal{L}_{\mathrm{rec}} =
\|\hat{X}_0 - X_0\|_2^2.
\end{equation}

While the reconstruction loss supervises individual joint positions, it does not explicitly preserve the structural relationships between connected joints. To encourage topology-consistent motion generation, we introduce a topology-aware bone-length loss.

Let $\mathcal{E}_K$ denote the edge set of the sparse skeleton. The bone-length loss is formulated as

\begin{equation}
\mathcal{L}_{\mathrm{bone}} =
\sum_{(i,j)\in \mathcal{E}_K}
\left(
\|\hat{x}_i-\hat{x}_j\|_2 -
\|x_i-x_j\|_2
\right)^2,
\end{equation}
where \(x_i\) and \(\hat{x}_i\) denote the ground-truth and predicted positions of joint \(i\), respectively. The temporal index is omitted for simplicity, and the loss is applied independently at each frame. This objective encourages the generated motion to preserve the structural characteristics of the underlying skeleton while remaining robust to variations in topology. Overall, the final training objective is
\begin{equation}
\mathcal{L}=
\mathcal{L}_{\mathrm{rec}}
+
\lambda_{\mathrm{bone}}
\mathcal{L}_{\mathrm{bone}}.
\end{equation}

%% file: sec/4_experiments.tex
\begin{figure*}[t]
    \centering
    \includegraphics[width=\textwidth]{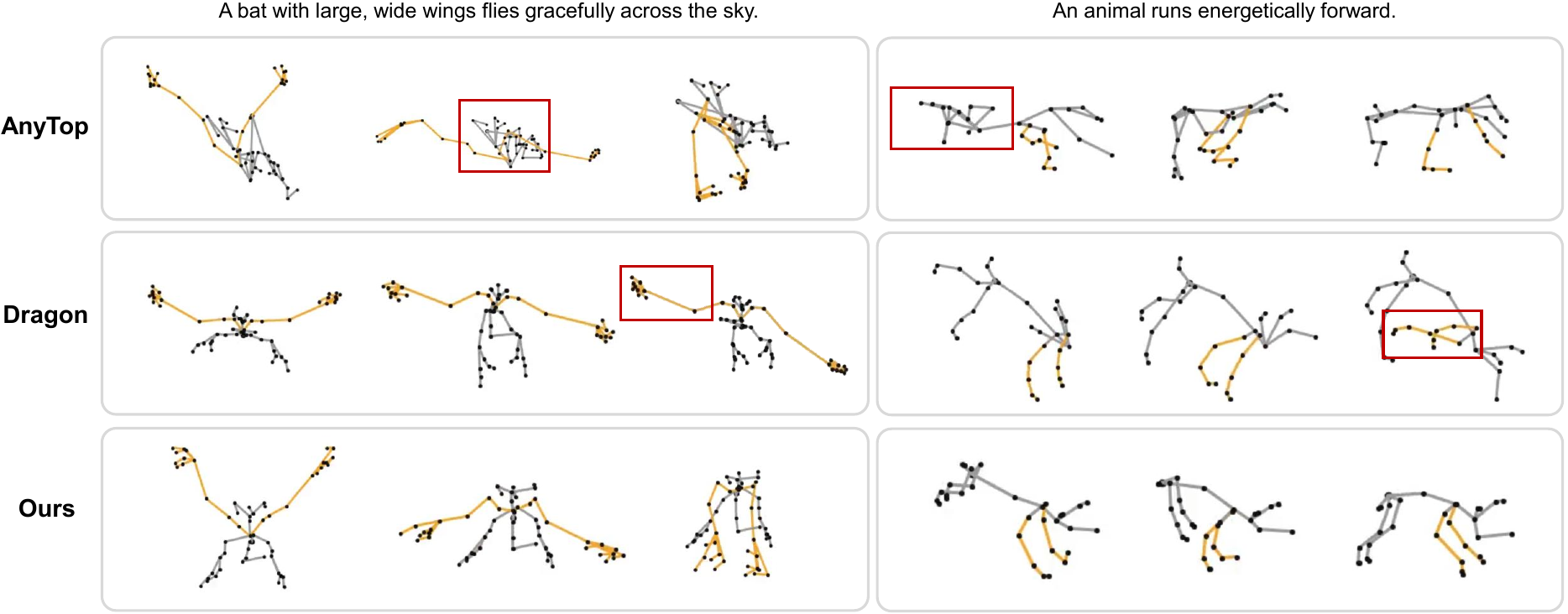}
    \caption{Qualitative comparison with baseline methods. The red boxes highlight representative failure cases of the baselines: topology distortions in AnyTop, including bone stretching and self-collision, and pose errors in Dragon that accumulate from inaccurate joint rotations. In contrast, ETA-Motion preserves a consistent skeletal structure while producing motions aligned with the text.}
    \label{fig:qualitative_results}
\end{figure*}

\section{Experiments}
\label{sec:experiments}

\subsection{Experimental Setup}
\paragraph{Dataset.}
We evaluate our approach on the Truebones Zoo dataset~\cite{truebones2022zoo}, which contains over 1,000 motion sequences from 74 species with diverse skeletal topologies. Since the dataset does not provide text annotations, we use the captions released with How to Move Your Dragon~\cite{Lee25Dragon} (hereafter, Dragon). We split the dataset into 65 training species and 9 test species. Following Dragon~\cite{Lee25Dragon}, we hold out one motion sequence from each training species for evaluation. We report results on three settings: (1) \textit{Seen}, where both the skeleton and motion are observed during training; (2) \textit{Unseen Motion}, where held-out motions from training species are used for evaluation; and (3) \textit{Unseen Topology}, where the model is tested on unseen species to measure generalization to novel skeletal topologies. Additional preprocessing details are provided in the supplementary material.

\begin{table*}[t]
    \centering
    \footnotesize
    \setlength{\tabcolsep}{4pt}
    \renewcommand{\arraystretch}{1.05}
    \caption{Quantitative comparison on the Seen, Unseen Motion, and Unseen Topology settings, together with training efficiency. AnyTop-G denotes the AnyTop variant trained with the shared global normalization scheme. Best results are highlighted in bold.}
    \label{tab:main_results}
    \resizebox{\textwidth}{!}{%
    \begin{tabular}{lcc ccc ccc ccc}
        \toprule
        \multirow{4}{*}{Method}
        & \multicolumn{2}{c}{Efficiency}
        & \multicolumn{3}{c}{Seen}
        & \multicolumn{3}{c}{Unseen Motion}
        & \multicolumn{3}{c}{Unseen Topology} \\
        \cmidrule(lr){2-3} \cmidrule(lr){4-6} \cmidrule(lr){7-9} \cmidrule(lr){10-12}
        & \makecell{GPU Mem.\\(MiB) $\downarrow$}
        & \makecell{Training\\(h) $\downarrow$}
        & Coverage (\%) $\uparrow$ & BLV $\downarrow$ & CR $\downarrow$
        & Coverage (\%) $\uparrow$ & BLV $\downarrow$ & CR $\downarrow$
        & Coverage (\%) $\uparrow$ & BLV $\downarrow$ & CR $\downarrow$ \\
        \midrule
        AnyTop
        & 21,180 & 66.44
        & 54.35 & 0.0710 & 0.0035
        & 42.24 & 0.5423 & 0.0039
        & 37.68 & 0.0624 & 0.0036 \\
        AnyTop-G
        & 21,507 & 67.80
        & 34.67 & 0.1419 & 0.0062
        & 24.01 & 0.1643 & 0.0063
        & 11.78 & 0.2392 & 0.0073 \\
        Dragon
        & 17,075 & 105.34
        & 72.66 & \textbf{0.0000} & 0.0030
        & 62.13 & \textbf{0.0000} & 0.0028
        & 66.82 & \textbf{0.0000} & 0.0043 \\
        Ours
        & \textbf{2,717} & \textbf{33.86}
        & \textbf{88.92} & 0.0012 & \textbf{0.0007}
        & \textbf{79.63} & 0.0009 & \textbf{0.0008}
        & \textbf{70.66} & 0.0010 & \textbf{0.0014} \\
        \bottomrule
    \end{tabular}%
    }
\end{table*}

\paragraph{Implementation Details.}
For the main baseline comparison, ETA-Motion is trained with a joint budget of \(K=20\). We provide additional analysis on the effect of varying \(K\) in the ablation study. Our model is based on a 4-layer DiT with latent dimension 128 and is trained with 100 diffusion steps. We train all models with a batch size of 16 on a single NVIDIA GeForce RTX 3090 or 4090 GPU. For the main baseline comparison, we report results at 600K training steps in consideration of the training budgets used in the baseline papers. Ablation studies are conducted at 300K steps to evaluate the relative effects of each design choice within our model.

\paragraph{Evaluation Metrics.}
Following Ganimator~\cite{li2022ganimator} and AnyTop~\cite{Gat25AnyTop}, we report \textit{Coverage} to measure how well generated motions reproduce motion patterns observed in the reference motion set. To evaluate topology preservation, we report two topology-aware metrics: \textit{Bone Length Variance (BLV)} and \textit{Capsule Collision Ratio (CR)}, where lower values indicate better structural consistency and fewer self-collisions. Since these automatic metrics do not directly evaluate semantic alignment, we additionally conduct a user study with 30 participants. Participants are asked to select the best result among Ours, AnyTop, and Dragon for two questions: \textit{skeleton-motion compatibility} and \textit{text-motion alignment}. We report the percentage of responses in which each method is selected as the best result. Additional details of the metric implementation and the user study are provided in the supplementary material.

\subsection{Comparison with Baselines}

\paragraph{Baselines.}
To the best of our knowledge, there is no publicly available implementation for text-conditioned motion generation on arbitrary skeletons. We therefore implement \textit{Dragon}~\cite{Lee25Dragon} and extend the official \textit{AnyTop}~\cite{Gat25AnyTop} implementation with text conditioning. Specifically, we implement Dragon based on the description in the paper, reproducing its key architectural components such as the tree embedding and rest-pose embedding. Since Dragon predicts 3D Euler joint rotations, we reconstruct joint positions via forward kinematics for evaluation. For AnyTop, we keep the official implementation unchanged and extend it only with text conditioning. Following the original AnyTop setup, this baseline applies joint-wise normalization using precomputed statistics for each joint of each asset. To enable a more direct comparison, we additionally train an AnyTop-G variant using the same global normalization scheme as ours, without joint-dependent statistics.

For all baselines, we preserve the model settings described in the original papers, including the latent dimension, number of attention heads, model depth, and batch size. For fair comparison, we standardize the T5 text encoder and the training dataset across all methods. Under this standardized setting, the implemented baselines do not match the performance reported in the original papers, which were trained with richer augmented data.

\paragraph{Quantitative Results.}
Table~\ref{tab:main_results} summarizes the overall quantitative comparison on the Seen, Unseen Motion, and Unseen Topology settings, together with training efficiency. Thanks to sparse joint training, ETA-Motion is substantially more GPU-efficient and requires less training time than the baselines. Beyond these efficiency gains, ETA-Motion also achieves the strongest overall performance across all three settings, with substantially higher coverage and lower CR than the baselines. Dragon yields zero BLV because its motions are reconstructed by forward kinematics with fixed bone lengths, whereas ETA-Motion achieves BLV values around 0.001 despite directly predicting joint positions.

\paragraph{User Study.}
As shown in Table~\ref{tab:user_study_main}, our method is preferred in 81.0\% of responses for \textit{skeleton-motion compatibility} and 74.8\% of responses for \textit{text-motion alignment}, substantially outperforming both AnyTop and Dragon. These results support that ETA-Motion better matches both the target skeleton and the input text than the baseline methods.

\begin{table}[h]
    \centering
    \small
    \setlength{\tabcolsep}{8pt}
    \renewcommand{\arraystretch}{0.95}
    \caption{User Preference on baseline comparison.}
    \label{tab:user_study_main}
    \begin{tabular}{lcc}
        \toprule
        Method & \makecell{Skeleton-Motion\\Compatibility (\%) $\uparrow$} & \makecell{Text-Motion\\Alignment (\%) $\uparrow$} \\
        \midrule
        AnyTop & 5.8 & 12.2 \\
        Dragon & 13.2 & 12.9 \\
        Ours & \textbf{81.0} & \textbf{74.8} \\
        \bottomrule
    \end{tabular}
\end{table}

\paragraph{Qualitative Results.}
Qualitative comparisons in Fig.~\ref{fig:qualitative_results} further highlight the distinct failure modes of the baselines. We additionally include an out-of-domain pig asset from Sketchfab~\cite{harris2021pig}, whose joints are assigned noisy labels. AnyTop often struggles to preserve the target topology, producing stretched limbs or bone intersections. For example, bones around the bat torso collide with each other, while the pig's legs become stretched and self-intersect. Dragon better preserves skeletal structure because motions are reconstructed through forward kinematics from predicted joint rotations. However, this formulation is sensitive to local rotation errors, which propagate along the kinematic chain and lead to large positional errors in descendant joints. As a result, the bat's hands fail to descend properly due to an incorrect shoulder rotation, and the pig's forelimb rotates excessively beyond a natural supporting pose. In contrast, ETA-Motion preserves skeletal consistency while producing motions that more faithfully match the text prompt. More extensive qualitative comparisons with baselines and mesh animation results are provided in the supplementary video.

\subsection{Ablation Studies}
\paragraph{Effect of Joint Budget.}

\begin{figure}[h]
    \centering
    \includegraphics[width=\columnwidth]{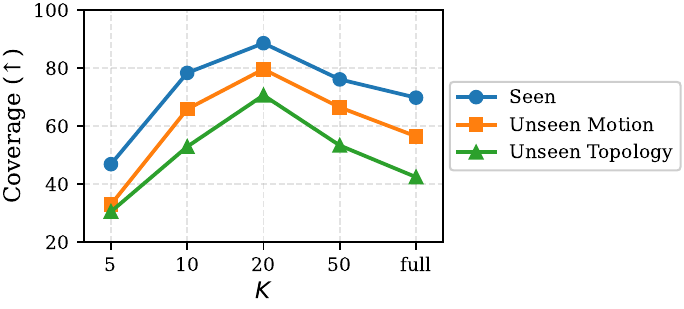}
    \caption{Coverage trends under different joint budgets \(K\).}
    \label{fig:k_budget}
\end{figure}

Table~\ref{tab:k_budget_efficiency} summarizes how the joint budget \(K\) affects efficiency, coverage, and human preference. Increasing \(K\) substantially raises computational cost, while generation quality improves only up to an intermediate budget. In particular, \(K=20\) achieves the best coverage across all evaluation settings and also obtains the highest Bradley--Terry preference scores for both skeleton-motion compatibility and text-motion alignment. Figure~\ref{fig:k_budget} further visualizes the coverage trends, showing clear gains from \(K=5\) to \(K=20\) followed by degradation at larger budgets. When \(K\) is too small, the sparse skeleton cannot capture sufficient articulated motion information, whereas overly large \(K\) increases modeling complexity with highly correlated joints. Overall, these results indicate that learning from a compact joint subset offers the best balance between efficiency and motion generation quality. More detailed results are provided in the supplementary material.

\begin{table*}[h]
    \centering
    \footnotesize
    \renewcommand{\arraystretch}{0.95}
    \caption{Efficiency, Coverage and Preference Score under different joint budgets \(K\). Best results are highlighted in bold.}
    \label{tab:k_budget_efficiency}
    \begin{tabular*}{\textwidth}{@{\extracolsep{\fill}}lccccccc}
        \toprule
        \multirow{3}{*}{$K$}
        & \multicolumn{2}{c}{Efficiency}
        & \multicolumn{3}{c}{Coverage (\%)}
        & \multicolumn{2}{c}{Preference Score (\%)} \\
        \cmidrule(lr){2-3} \cmidrule(lr){4-6} \cmidrule(lr){7-8}
        & \makecell{GPU Mem.\\(MiB) $\downarrow$}
        & \makecell{Training\\(h) $\downarrow$}
        & \makecell{Seen $\uparrow$}
        & \makecell{Unseen \\Motion $\uparrow$}
        & \makecell{Unseen \\Topology $\uparrow$}
        & \makecell{Skeleton\\Compatibility $\uparrow$}
        & \makecell{Text\\Alignment $\uparrow$} \\
        \midrule
        5    & 1,749  & 16.10 & 44.16 & 32.99 & 30.54 & 2.0  & 4.0 \\
        10   & 1,967  & 16.06 & 77.80 & 65.87 & 52.94 & 7.1  & 13.7 \\
        20   & 2,717  & 16.84 & \textbf{88.60} & \textbf{75.57} & \textbf{65.73} & \textbf{58.1} & \textbf{37.8} \\
        50   & 4,941  & 19.09 & 74.06 & 66.49 & 53.36 & 24.3 & 31.8 \\
        full & 12,688 & 27.44 & 67.46 & 56.50 & 42.45 & 8.6  & 12.6 \\
        \bottomrule
    \end{tabular*}
\end{table*}

\paragraph{Component Ablations.}
Table~\ref{tab:component_ablation_main} evaluates the effects of the visual joint embedding and bone-length loss. Overall, both components are beneficial: the visual embedding improves coverage in the Seen and Unseen Motion settings, while the bone-length loss consistently improves performance across all evaluation settings. On Unseen Topology, visual embeddings yield slightly lower coverage because clean joint labels already provide strong semantic cues. Their robustness to noisy labels is illustrated in the supplementary qualitative results.

We compare the position-only and redundant representations in the full-joint setting. This is because, under sparse joint training, rotations must be recomputed from the selected joints, which introduces additional noise and makes the comparison less direct. In this controlled setting, the position-only representation achieves higher coverage on both unseen motion and unseen topology. More detailed ablation results are provided in the supplementary material.

\begin{table}[h]
    \centering
    \scriptsize
    \renewcommand{\arraystretch}{1.05}
    \caption{Component ablations on the Seen, Unseen Motion, and Unseen Topology settings in terms of coverage (\%). Best results are highlighted in bold.}
    \label{tab:component_ablation_main}
    \resizebox{\columnwidth}{!}{%
    \begin{tabular}{llccc}
        \toprule
        K & Method & Seen $\uparrow$ & \makecell{Unseen\\Motion $\uparrow$} & \makecell{Unseen\\Topology $\uparrow$} \\
        \midrule
        \multirow{3}{*}{20}
        & Ours            & \textbf{88.60} & \textbf{75.57} & 65.73 \\
        & w/o visual emb. & 82.52 & 73.15 & \textbf{68.41} \\
        & w/o bone loss   & 77.52 & 67.50 & 63.74 \\
        \midrule
        \multirow{2}{*}{Full}
        & Ours              & \textbf{67.46} & \textbf{56.50} & \textbf{42.45} \\
        & w/ redundant rep. & 50.34 & 48.03 & 35.29 \\
        \bottomrule
    \end{tabular}%
    }
\end{table}

%% file: sec/5_conclusion.tex
\section{Conclusion}
\label{sec:conclusion}
In this paper, we present ETA-Motion, an efficient topology-agnostic motion generation framework based on sparse joint learning. Unlike prior approaches that rely on dense full-joint motion modeling, ETA-Motion learns motion dynamics from a compact joint subset while generating motions for full target skeletons at inference time. To support transferable motion generation across diverse skeletal topologies, we introduce semantic joint embeddings and topology-aware training objectives. Experimental results show that ETA-Motion consistently improves both motion quality and training efficiency over baselines. Ablation studies further confirm that the performance gains arise from the proposed sparse joint design, semantic joint embedding, topology-aware supervision, and compact position-based motion representation. These results suggest that compact sparse motion representations provide an effective alternative to dense full-joint modeling for topology-agnostic motion generation.

\paragraph{Limitations and Future Works.} Despite its strong efficiency and generation performance, ETA-Motion has several limitations. Our current implementation uses FPS as a simple heuristic for constructing sparse joint subsets. Although effective in practice, it may not always retain the most informative joints. In addition, because our model predicts joint positions, practical deployment to rigged meshes requires an additional inverse kinematics stage. This additional stage often leads to temporal inconsistency or reconstruction artifacts. Future work includes integrating a dedicated position-to-rotation module, as explored in recent studies~\cite{gong2026mocapanything, Gong26MoCapAnythingV2}, to further improve practical usability.

\section*{Acknowledgments}
This work was supported by NST grant (CRC 21015, MSIT), IITP grant (RS-2023-00228996, RS-2024-00459749, RS-2025-25443318, RS-2025-25441313, RS-2026-25526850, RS-2026-25522885, MSIT), KOCCA grant (RS-2024-00442308, MCST) and InnoCORE program (N10260110, MSIT).

\par